\documentclass{article}

\usepackage{microtype}
\usepackage{graphicx}
\usepackage{subcaption}
\usepackage{booktabs} % for professional tables
\usepackage{multirow}
\usepackage{array}

\usepackage{hyperref}

\usepackage[preprint]{icml2026_arxiv}

\usepackage{amsmath}
\usepackage{amssymb}
\usepackage{mathtools}
\usepackage{amsthm}

\usepackage[capitalize,noabbrev]{cleveref}

\theoremstyle{plain}

\theoremstyle{definition}

\theoremstyle{remark}

\usepackage[textsize=tiny]{todonotes}

\usepackage{needspace}

\icmltitlerunning{Training-free Dropout Sampling for Semantic Token Acceptance in Speculative Decoding}

\makeatletter
\renewcommand{\@pa}[1]{%
  \ifcsname the@affil#1\endcsname
  \else
    \ifcsname @icmlsymbol#1\endcsname
    \else
      \stepcounter{@affiliationcounter}%
      \newcounter{@affil#1}%
      \setcounter{@affil#1}{\value{@affiliationcounter}}%
    \fi
  \fi%
  \ifcsname @icmlsymbol#1\endcsname
    \textsuperscript{\csname @icmlsymbol#1\endcsname\,}%
  \else
  \fi
}
\makeatother

\makeatletter
\renewcommand{\Notice@String}{}
\renewcommand{\printAffiliationsAndNotice}[1]{%
  \ificml@noticeprinted\relax
  \else
    \global\icml@noticeprintedtrue
    \begin{center}
      \footnotesize
      NAVER Cloud, SeongNam-si, South Korea\\
      \texttt{\{jeongtae.lee,dongsoo.lee\}@navercorp.com}
    \end{center}
  \fi
}
\makeatother

\begin{document}

\twocolumn[
  \icmltitle{Training-free Dropout Sampling for Semantic \\
  Token Acceptance in Speculative Decoding}

  \icmlsetsymbol{equal}{*}
  
  \begin{icmlauthorlist}
    \icmlauthor{Jeongtae Lee}{navercloud}
    \icmlauthor{Minjung Jo}{navercloud}
    \icmlauthor{Hyunjoon Jeong}{navercloud}
    \icmlauthor{Gunho Park}{navercloud}
    \icmlauthor{Sunghyeon Woo}{navercloud}
    \icmlauthor{Joonghoon Kim}{navercloud}
    \icmlauthor{Se Jung Kwon}{navercloud}
    \icmlauthor{Dongsoo Lee}{navercloud}
  \end{icmlauthorlist}

  %\icmlaffiliation{navercloud}{Naver Cloud, SeongNam-si, South Korea}
  %\icmlcorrespondingauthor{Jeongtae Lee}{jeongtae.lee@navercorp.com}
  \printAffiliationsAndNotice{}
  \vskip 0.3in
]

% this must go after the closing bracket ] following \twocolumn[ ...

% This command actually creates the footnote in the first column listing the
% affiliations and the copyright notice. The command takes one argument, which
% is text to display at the start of the footnote. The \icmlEqualContribution
% command is standard text for equal contribution. Remove it (just {}) if you
% do not need this facility.

% Use ONE of the following lines. DO NOT remove the command.
% If you have no special notice, KEEP empty braces:
\printAffiliationsAndNotice{}  % no special notice (required even if empty)
% Or, if applicable, use the standard equal contribution text:
% \printAffiliationsAndNotice{\icmlEqualContribution}

\begin{abstract}
   %We propose a novel speculative decoding method for large language models (LLMs) that applies Monte Carlo (MC) dropout to the LM head to enable sampling-based acceptance decisions. Our approach applies MC dropout exclusively to the target model’s LM head, performing inference as if multiple heads were present and thereby enabling effective token sampling. Tokens proposed by a draft model are evaluated based on a distribution formed by multiple MC dropout predictions. Draft tokens are accepted when they are sufficiently consistent with the target model’s predictive distribution. This acceptance mechanism flexibly adapts to different distributional shapes, allowing stable handling of diverse sampling behaviors. The proposed method operates in a training-free, calibration-free, and data-free manner, requiring only a pretrained model. Experiments on various datasets show that our approach effectively increases acceptance length while maintaining competitive task performance. Furthermore, we demonstrate that the proposed method is orthogonal to a wide range of speculative decoding and inference acceleration techniques.
   
Speculative decoding accelerates large language model inference by proposing tokens with a lightweight draft model and selectively accepting them using a target model. This work introduces DropMatch, a novel approach that matches draft tokens to the predictive distribution of the target model via Monte Carlo dropout applied exclusively to the LM head, enabling sampling-based acceptance decisions. By generating multiple decoding paths, our method forms an empirical token distribution against which draft tokens are evaluated for consistency. This acceptance mechanism enables the model to adaptively control the size of decoding paths under an appropriate dropout probability, preventing substantial distortion of the target model predictive distribution. The proposed method operates in a training-free, data-free, and calibration-free manner, requires no architectural modification to pretrained models, and can be orthogonally integrated with a wide range of existing speculative decoding and inference acceleration techniques. Experiments across multiple benchmarks demonstrate that our approach increases acceptance length while maintaining competitive task performance, yielding inference speedups ranging from 1.09× to 1.33× over the standard baseline, and up to an additional 1.09x speedup when applied on top of EAGLE3.
  
\end{abstract}

\section{Introduction} \label{introduction}

Large language models (LLMs) have consistently demonstrated improved performance across a wide range of tasks as model size and computational capacity increase~\cite{grattafiori2024llama3herdmodels, qwen3, kaplan2020scalinglawsneurallanguage}. 
Such scaling trends have been observed across diverse domains, including language understanding, reasoning~\cite{wei2023chainofthoughtpromptingelicitsreasoning}, and code generation~\cite{rozière2024codellamaopenfoundation}, indicating that increases in model scale lead to stronger representational and generalization capabilities.
However, such gains inevitably come with increased inference costs, making efficient inference with preserved model performance a central challenge for practical deployment.
One of the primary bottlenecks in LLM inference lies in the auto-regressive decoding process~\cite{shazeer2019fasttransformerdecodingwritehead}.
Under auto-regressive decoding, tokens are generated sequentially, with each token conditioned on all previously generated tokens, enforcing a strictly serial computation pattern.
For example, DeepSeek-R1~\cite{deepseek_r1, deepseek-ai_deepseek-v3_2024}, which has 671 billion total parameters, typically requires partitioning across multiple NVIDIA H200 GPUs to fit the model weights in memory, yet still takes several seconds to process a single inference request.
Moreover, inference latency becomes even more amplified in reasoning mode~\cite{openai2025gptoss} or agentic workloads~\cite{yao2023react}, where the number of generated tokens increases substantially.
Consequently, even with abundant computational resources, the benefits of parallelism are difficult to fully exploit during decoding. This strict sequential imposed by auto-regressive generation has been widely recognized as a major source of inference inefficiency.

To address this limitation, various inference acceleration techniques such as speculative decoding have been proposed~\cite{pmlr-v202-leviathan23a, chen2023acceleratinglargelanguagemodel}. Speculative decoding improves decoding efficiency by allowing a smaller draft model to propose multiple tokens in advance, which are then verified by a larger target model. A key factor in speculative decoding is how many proposed tokens can be accepted in a single verification step, as the acceptance length directly determines the overall speedup. Therefore, increasing acceptance length without sacrificing accuracy remains a central challenge in speculative decoding.

In this work, we propose DropMatch, a sampling-based acceptance method that leverages Monte Carlo (MC) dropout~\cite{pmlr-v48-gal16} to improve the efficiency of speculative decoding.
Our approach applies MC dropout exclusively to the LM head of the target model, allowing multiple stochastic forward passes at the head level to produce diverse token samples without additional training or calibration. 
The resulting MC dropout samples are used to efficiently evaluate whether draft tokens are consistent with the target model predictions, increasing acceptance length with minimal computational overhead. 
As a result, DropMatch achieves practical speedups without modifying the overall speculative decoding framework.

We demonstrate the effectiveness of DropMatch across multiple model families~\cite{grattafiori2024llama, yang2025qwen3} and a broad set of reasoning~\cite{cobbe2021gsm8k}, language understanding~\cite{hendryckstest2021}, and instruction following~\cite{zhou2023instructionfollowingevaluationlargelanguage} benchmarks. 
Overall, DropMatch consistently improves acceptance length and translates these acceptance gains into end to end decoding speedups while preserving task performance. 
In addition, DropMatch integrates seamlessly with recent speculative decoding pipelines~\cite{li2025eagle3}, enabling further improvements when combined with additional drafting and verification components. Finally, when paired with an external judging mechanism~\cite{garipov2025autojudgejudgedecodingmanual}, DropMatch enables a practical and tunable accuracy-latency trade-off, allowing users to adjust conservativeness while retaining measurable speed benefits.

In summary, our contributions are as follows:
\begin{itemize}
  \item We introduce DropMatch, a sampling based acceptance method for speculative decoding that leverages MC dropout applied only to the target model LM head. This design yields multiple token samples within a single decoding step, avoiding temperature-based heuristics and repeated full model evaluations.
  \item DropMatch requires no training, calibration, or auxiliary data, and adds only a small additional verification cost. It consistently increases acceptance length and translates these gains into end to end decoding speedups, yielding improved accuracy latency trade offs.
  \item We demonstrate that DropMatch is compatible with a wide range of speculative decoding and inference acceleration techniques. As a result, it can be seamlessly combined with existing methods to further improve decoding efficiency without sacrificing their original advantages.
\end{itemize}

\section{Related works} \label{Related-works}

\paragraph{Speculative Decoding} Speculative decoding~\cite{pmlr-v202-leviathan23a, xia2023speculativedecodingexploitingspeculative} is a representative acceleration technique designed to improve the inference throughput of large target models by verifying tokens proposed by a smaller draft model. In a typical setup, the draft model first generates a sequence of tokens of fixed length, and the target model performs a single verification step to determine which of these tokens can be accepted. The number of accepted tokens directly reduces the number of auto-regressive decoding steps required by the target model, leading to faster inference. As a result, better alignment~\cite{zhou2024distillspec,hu2025griffin} between the draft and target models generally leads to a higher average number of accepted tokens and greater acceleration.

\paragraph{Lossless Speculative Decoding} Lossless speculative decoding guarantees that the final output tokens follow exactly the same sampling distribution as the target model through a strict verification process. Prior work~\cite{li2025eagle3, cai2024medusa} in this category has focused on strengthening the alignment between draft and target models while enabling faster token proposal and verification. Additionally, since a rejection at a certain position causes all subsequent tokens to be rejected, some studies~\cite{li2025gumihohybridarchitectureprioritize, huang2025posspositionspecialistgenerates, zhang2025learning} have proposed using depth-specialized models to improve acceptance rates. However, in lossless approaches, even semantically equivalent tokens are rejected if they differ at the token level, which fundamentally limits the achievable speedup.

\begin{figure}[h]
  \centering
  \begin{subfigure}[t]{0.48\columnwidth}
    \centering
    \includegraphics[width=\linewidth]{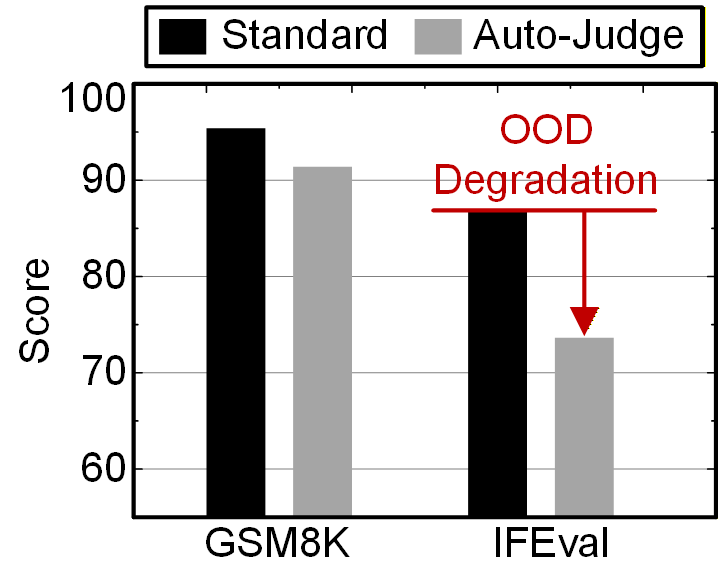}
    \caption{Trained on math data}
    \label{fig:ifeval_ood:a}
  \end{subfigure}
  % \hfill
  \begin{subfigure}[t]{0.48\columnwidth}
    \centering
    \includegraphics[width=\linewidth]{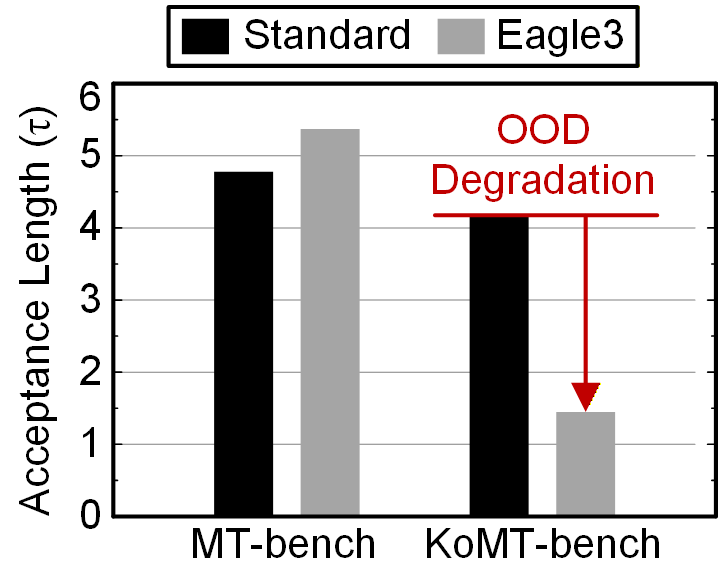}
    \caption{Trained on English data}
    \label{fig:komt_ood:b}
  \end{subfigure}
  \caption{
    Out-of-distribution performance of Auto-Judge on IFEval and EAGLE3 on KoMT-Bench. (a) Results of Auto-Judge using a judge head trained with the Llama-3.1-70B-Instruct model under out-of-distribution conditions, exhibiting preserved acceptance length but degraded task performance.
    (b) Results of EAGLE3 using a draft model trained with the Llama-3.3-70B-Instruct model, demonstrating that task performance is maintained while acceptance length decreases on KoMT-Bench, leading to reduced acceleration benefits. $\tau$ denotes the mean acceptance length.
  }
  \label{ifeval_komt_ood}
\end{figure}

\paragraph{Lossy Speculative Decoding}
Lossy speculative decoding relaxes the strict requirement of sampling from the exact distribution of the target model, allowing tokens proposed by a draft model to be accepted as long as the output distribution is preserved. 
A representative approach is Judge Decoding~\cite{bachmann2025judge}, which trains a dedicated judge head from human annotations to detect positions where a token substitution would change semantics. 
More recent methods reduce or remove human supervision: Auto-Judge~\cite{garipov2025autojudgejudgedecodingmanual} and Self-Judge~\cite{yoon2025selfjudgefasterspeculativedecoding} let models identify quality-critical positions autonomously, but still rely on additional learned components or auxiliary training data. 
This reliance can limit robustness under domain shift: when the training distribution of the judge head or the draft model is narrow, effectiveness can degrade on out-of-distribution benchmarks. 
\cref{ifeval_komt_ood} illustrates this effect empirically: Auto-Judge, where the judge head is trained on mathematical data, shows reduced performance on IFEval, and EAGLE3, where the draft model is trained on English data, exhibits markedly shorter acceptance lengths on the Korean KoMT-bench~\cite{KoMT-Bench} benchmark.

% \begin{table}[h]
% \centering
% \caption{Out-of-distribution performance of Auto-Judge on IFEval with the Llama-3.1-70B-Instruct model and of EAGLE3 on KoMT-Bench with the Llama-3.3-70B-Instruct model. The judge head of Auto-Judge is trained on mathematical data, and the draft model of EAGLE3 is trained exclusively on English data. $\tau$ represents the average acceptance length.}
% \label{tab:ifeval_komt}
% \setlength{\tabcolsep}{6pt}
% \small
% \begin{tabular}{lcc||lc}
% \toprule
% \multicolumn{3}{c||}{\textbf{IFEval}} & \multicolumn{2}{c}{\textbf{KoMT-bench}} \\
% \midrule
% Model & Score & $\tau$ & Model & $\tau$ \\
% \midrule
% Standard        & 86.86 & 5.47 & Standard    & 4.17 \\
% Auto-Judge & 73.63 & 25.67 & EAGLE3 & 1.45 \\
% \bottomrule
% \end{tabular}
% \end{table}

The proposed method, DropMatch, falls into the category of lossy speculative decoding. When tokens are semantically similar, sampling is likely to produce overlapping or closely aligned token candidates. Even when identical tokens are not generated, sufficiently similar probability distributions can be considered to follow the distribution of the target model. Our approach does not require training new architectures such as EAGLE~\cite{li2024eagle} or POSS~\cite{huang2025posspositionspecialistgenerates}, nor does it rely on additional data or judge heads as in Judge Decoding~\cite{bachmann2025judge} or Auto-Judge~\cite{garipov2025autojudgejudgedecodingmanual}. Furthermore, the proposed method operates without any calibration process~\cite{gautam2025tokendrivengammatuneadaptivecalibration}, offering a simple and broadly applicable extension to speculative decoding.

\begin{figure}[H]
  \vskip 0.2in
  \begin{center}
    \centerline{\includegraphics[width=7cm]{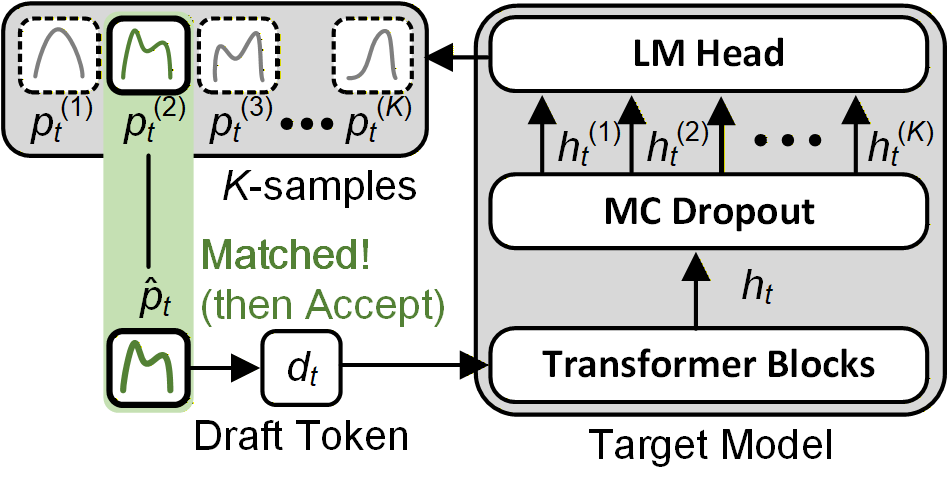}}
    \caption{
      Overall architecture of DropMatch, illustrating speculative decoding with multiple sampling enabled by MC dropout applied at the LM head. $d_t$ denotes the $t$-th draft token, and $h_t$ denotes its corresponding final embedding vector. $h_t^{(i)},\dots,h_t^{(K)}$ represent $K$ MC dropout paths generated by applying $K$ different dropout masks to the $t$-th embedding.
    }
    \label{fig:archtecture}
  \end{center}
\end{figure}

\section{Methodology}
In this section, we describe how DropMatch enables efficient token sampling and how the resulting samples are used to determine token acceptance. Prior speculative decoding methods are typically inductive, relying on training the draft model to closely approximate the output distribution of the target model. In contrast, our approach operates in a transductive manner at inference time, directly leveraging the predictive distribution of the target model, which can be interpreted similarly to a k-nearest neighbor~\cite{journals/tit/CoverH67} mechanism.

% \subsection{MC dropout LM Head} \label{MCdrop_lmhead}
\subsection{Multi-Sample LM Head via MC Dropout} \label{MCdrop_lmhead}

MC dropout has been widely used to approximate ensemble effect~\cite{srivastava2014dropout} and to quantify predictive uncertainty~\cite{pmlr-v48-gal16} in neural networks. 
We revisit MC dropout as a sampling mechanism and propose a novel method, termed DropMatch, to assess whether tokens proposed by a draft model are consistent with predictions of the target model. 
To avoid repeated full forward passes and excessive computation, MC dropout is applied exclusively to the LM head rather than to the entire network. 
By applying MC dropout only at the LM head, this design preserves KV-cache alignment of the remaining transformer blocks, making it straightforward to implement.

For clarity, the LM head is modeled as producing $K$ stochastic predictions by applying independent dropout masks to the final hidden representation. 
Let $h_t \in \mathbb{R}^d$ denote the last-layer hidden state at time step $t$. 
For each path $i \in \{1,\dots,K\}$, a mask $m^{(i)} \in \{0,1\}^d$ is sampled with i.i.d.\ entries:
\begin{equation}\label{eq:mask}
m^{(i)}_j \overset{\text{i.i.d.}}{\sim} \mathrm{Bernoulli}(1 - p_{\text{drop}}),
\quad j = 1, \dots, d .
\end{equation}
Using inverted-dropout scaling, the masked representation is defined as
\begin{equation}\label{eq:dropout}
h_t^{(i)} = \frac{h_t \odot m^{(i)}}{1 - p_{\text{drop}}}, \quad i = 1, \dots, K .
\end{equation}
Eq.~\eqref{eq:dropout} preserves the expectation $\mathbb{E}[h_t^{(i)}]=h_t$, matching the standard dropout convention. 
The LM head then produces the corresponding logits $l_t^{(i)}$ and token probabilities $p_t^{(i)}$ as
\begin{equation}\label{eq:logits_softmax}
l_t^{(i)} = W h_t^{(i)}, \quad
p_t^{(i)} = \mathrm{Softmax}\!\left(l_t^{(i)}\right), \quad i = 1, \dots, K .
\end{equation}
All paths share the same LM-head weights $W$ and differ only through the sampled dropout masks, enabling $K$ parallel samples without introducing additional parameters. \cref{fig:archtecture} illustrates the overall architecture of the proposed method.

\begin{figure*}[!t]
  \centering
  % ---------- (a) ----------
  \begin{minipage}[t]{0.49\textwidth}
    \centering
    \begin{minipage}{0.49\linewidth}
      \centering
      \includegraphics[width=\linewidth]{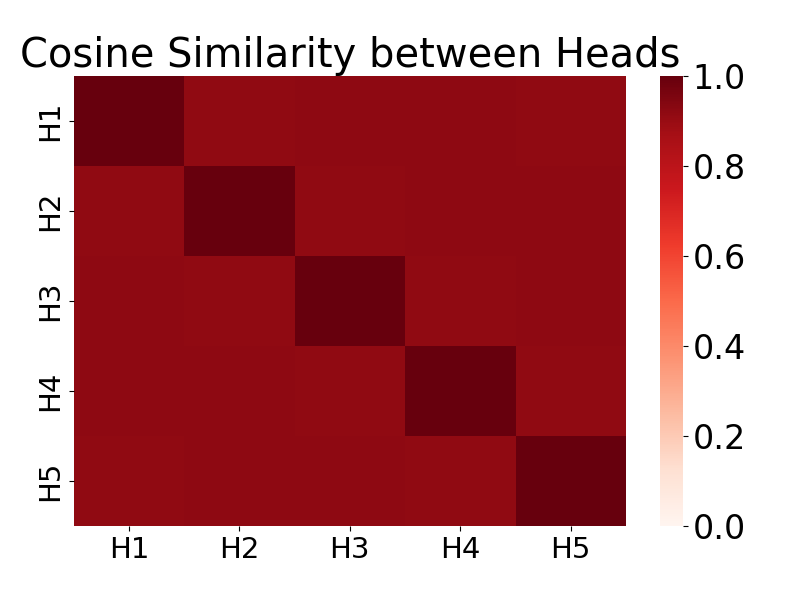}
    \end{minipage}
    %\hspace{0.001\textwidth}
    \hspace{-6pt}
    \begin{minipage}{0.49\linewidth}
      \centering
      \includegraphics[width=\linewidth]{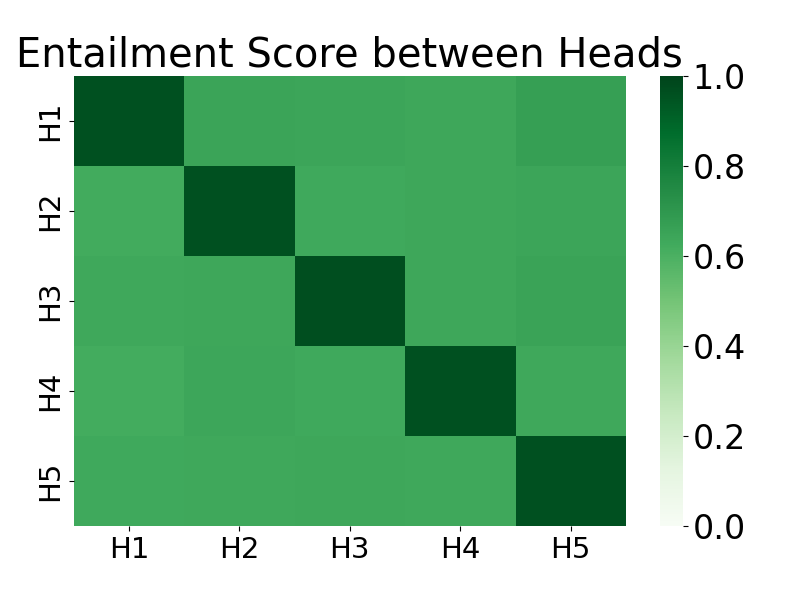}
    \end{minipage}
    \vspace{3pt}
    \par\centering \textbf{(a)} dropout probability $p_{drop}=0.1$
  \end{minipage}
  \hspace{0.01\textwidth}
  % ---------- (b) ----------
  \begin{minipage}[t]{0.49\textwidth}
    \centering
    \begin{minipage}{0.49\linewidth}
      \centering
      \includegraphics[width=\linewidth]{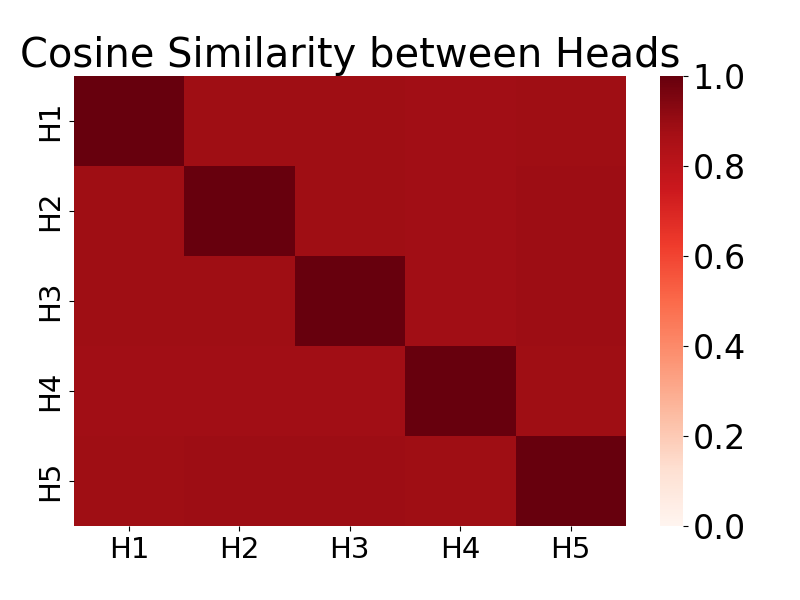}
    \end{minipage}
    %\hspace{0.001\textwidth}
    \hspace{-6pt}
    \begin{minipage}{0.49\linewidth}
      \centering
      \includegraphics[width=\linewidth]{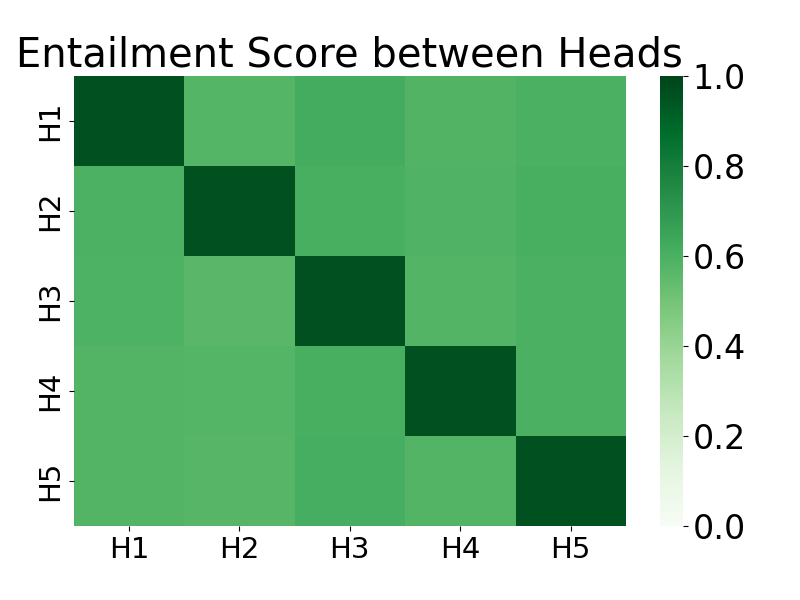}
    \end{minipage}
    \vspace{3pt}
    \par\centering \textbf{(b)} dropout probability $p_{drop}=0.3$
  \end{minipage}
  \caption{Semantic similarity across multiple decoding paths. (a) Cosine similarity matrices computed with Sentence-BERT and semantic consistency matrices from a sentence entailment model at dropout probability $p_{drop}=0.1$, (b) Corresponding results at $p_{drop}=0.3$, showing that lower dropout probabilities yield higher semantic similarity across paths. H1--H5 denote the MC dropout with $K=5$ decoding paths, each corresponding to a distinct stochastic forward pass through the LM head. The higher the value, the darker the color.}
  \label{sim_matrix}
\end{figure*}

As a preliminary validation, we provide empirical evidence that multiple decoding paths generated by MC dropout produce semantically consistent outputs using the Llama-3.1-70B-Instruct~\cite{grattafiori2024llama3herdmodels} model on the spec-bench~\cite{xia-etal-2024-unlocking} dataset. 
\cref{sim_matrix} follows the semantic uncertainty~\cite{kuhn2023semanticuncertaintylinguisticinvariances} evaluation protocol, measuring cosine similarity between token sequences generated by different paths using sentence-BERT~\cite{reimers-2019-sentence-bert}, as well as semantic consistency using a sentence entailment model~\cite{he2021deberta}. We use $K=5$ paths and observe higher semantic similarity at lower dropout probabilities, indicating that the LM head outputs remain semantically aligned even without explicit MC dropout training.
\cref{each_head_perf} further reports the performance of individual heads on the HumanEval benchmark~\cite{chen2021evaluating}. The results show that each path produced by the LM head continues to outperform the draft model and maintains comparable accuracy when the dropout probability is moderate. Based on these observations, we demonstrate in \cref{Experiments} that aligning with at least one of multiple sampled paths during the verification phase can effectively increase acceptance length without degrading task performance.

\begin{table}[h]
    \centering
    \small
    \setlength{\tabcolsep}{2.5pt}
    \caption{Pass@1 performance of Llama-3.1-70B-Instruct across different heads and dropout probabilities. Baseline Pass@1 of 81.7 without dropout; Llama-3.1-8B-Instruct used as the draft model with Pass@1 of 72.6.}
    \label{each_head_perf}
    \begin{tabular}{cccccc}
    \toprule
    \textbf{L31 70B Inst} & \textbf{Head1} & \textbf{Head2} & \textbf{Head3} & \textbf{Head4} & \textbf{Head5} \\
    \midrule
    Pass@1 ($p{=}0.1$) & 81.1 & 78.7 & 81.1 & 86.0 & 77.4 \\
    Pass@1 ($p{=}0.2$) & 80.5 & 81.1 & 80.5 & 81.1 & 78.7 \\
    Pass@1 ($p{=}0.3$) & 78.7 & 81.1 & 75.6 & 77.4 & 79.9 \\
    Pass@1 ($p{=}0.4$) & 76.2 & 79.3 & 80.5 & 73.2 & 76.2 \\
    Pass@1 ($p{=}0.5$) & 78.7 & 65.2 & 73.8  & 64.0 & 70.1 \\
    \bottomrule
    \end{tabular}
\end{table}

\subsection{Acceptance Criteria} \label{accept_criteria}
Standard speculative decoding determines token acceptance based on rejection sampling by comparing token probabilities from the draft and target models. In our approach, we treat the $K$ output distributions from the multiple decoding paths as a single cluster and evaluate whether the draft token belongs to this cluster.

\paragraph{Naive Token-Matching Criterion}
The simplest criterion accepts a draft token if it matches any token produced by the $K$ decoding paths.
Let $\hat{y}_t$ denote the token proposed by the draft model at time step $t$, and let
$y_t^{(i)}=\arg\max\!\big(p_t^{(i)}\big)$ denote the token selected by the $i$-th head.
The acceptance condition is
\begin{equation}\label{eq3}
    \hat{y}_t \in \{ y_t^{(i)} \mid i=1,\ldots,K \}.
\end{equation}
When the dropout heads yield sufficiently stable predictions, Eq.~\ref{eq3} can substantially increase the acceptance length with negligible overhead.
However, because this rule ignores the full probability mass and relies only on the top-1 tokens or selected token by nucleus sampling~\cite{holtzman2020curiouscaseneuraltext}, it can accept a draft token even when the underlying distributions are not well aligned.

\begin{figure}[h]
  \centering
  \begin{subfigure}[t]{0.48\columnwidth}
    \centering
    \includegraphics[width=\linewidth]{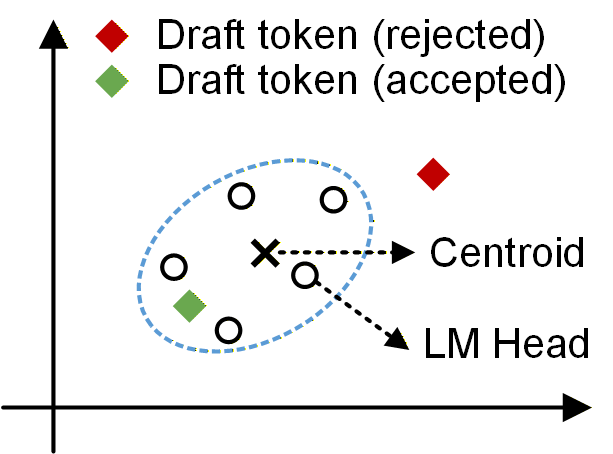}
    \caption{Dispersed samples}
    \label{fig:cluster:a}
  \end{subfigure}
  % \hfill
  \begin{subfigure}[t]{0.48\columnwidth}
    \centering
    \includegraphics[width=\linewidth]{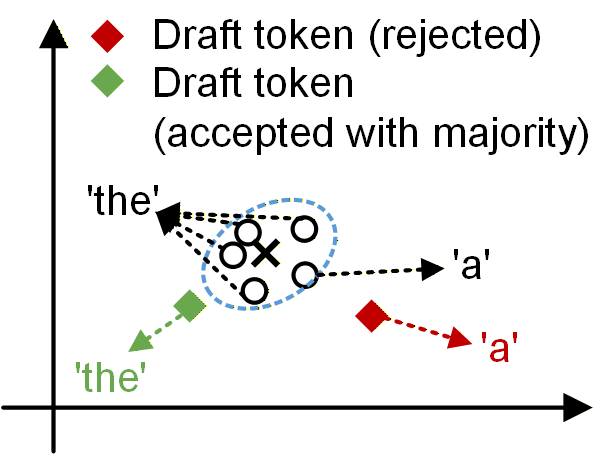}
    \caption{Concentrated samples}
    \label{fig:cluster:b}
  \end{subfigure}
  \caption{
    Conceptual illustration of the JS-divergence–based acceptance criterion. (a) Acceptance determined solely by Eq.~\ref{eq5} under dispersed MC dropout sample distributions. (b) Acceptance determined by Eq.~\ref{eq6} under highly concentrated sample distributions. Both subfigures illustrate acceptance and rejection cases.
  }
  \label{mc_drop_dist_fig}
\end{figure}

\paragraph{JS-Divergence--Based Criterion}
To address this limitation, we more generally compare the draft distribution with the cluster of MC dropout head distributions using Jensen--Shannon (JS) divergence.
We first define a centroid distribution by averaging the head logits and normalizing:
\begin{equation}\label{eq4}
    \bar{p}_t = \mathrm{Softmax}\!\left(\frac{1}{K}\sum_{i=1}^{K} l_t^{(i)}\right),
\end{equation}
where $l_t^{(i)}$ denotes the logits produced by the $i$-th head.
Let $\hat{p}_t$ denote the draft model distribution at time step $t$.
We accept the draft token if its divergence to the centroid is no larger than the maximum divergence observed among the MC dropout heads:
\begin{equation}\label{eq5}
    \mathrm{JS}\!\left(\hat{p}_t \,\|\, \bar{p}_t\right)
    \;\le\;
    \max_{i=1,\ldots,K}
    \mathrm{JS}\!\left(p_t^{(i)} \,\|\, \bar{p}_t\right).
\end{equation}

This criterion evaluates whether the draft token lies within the sampling distribution of the target model.
When the samples from each LM head distributions are highly similar, the draft distribution may still be rejected despite being sufficiently close (Fig. \ref{fig:cluster:b}). Since such cases indicate that the target model has effectively collapsed to a single dominant output, we introduce an additional criterion:
\begin{equation}\label{eq6}
    \mathrm{majority}\left(\{ y_t^{(i)} \mid i=1,\ldots,K \} \right) = \hat{y}_t
\end{equation}
In other words, a draft token is accepted if it matches the majority token among the multiple heads. \cref{tab:agreement_stats} reports how frequently majority tokens occur under Eq.~\ref{eq6} on the HumanEval dataset. With $K=5$, all heads predict the same token in 98.4\% of cases, if heads are not dispersed, which is substantially higher than other outcomes, indicating that a strong majority is common on HumanEval. \cref{tab:js_agreement} further compares the average JS divergence values for cases accepted under Eq.~\ref{eq5} and those accepted under Eq.~\ref{eq6}. Notably, cases accepted by Eq.~\ref{eq6} often exhibit very low JS divergence, yet would be rejected when relying solely on Eq.~\ref{eq5}, highlighting the limitation of using the JS-divergence criterion alone. The complete acceptance procedure is summarized in \cref{accept_algo}.

\begin{table}[t]
\centering
\caption{Statistics of head alignment and the corresponding mean probabilities conditioned on alignment for the Llama-3.1-70B-Instruct model evaluated on HumanEval with MC dropout using $K=5$ heads.}
\label{tab:agreement_stats}
\setlength{\tabcolsep}{4pt}
\small
\begin{tabular}{p{1.4cm} cccccc}
\toprule
 Majority & 1 & 2 & 3 & 4 & 5 \\
\midrule
Ratio   & 0.00\%     & 0.14\%    & 0.82\%   & 0.54\%   & 98.40\% \\
Mean prob. & 0.255 & 0.325 & 0.460 & 0.517 & 0.939 \\
\bottomrule
\end{tabular}
\end{table}

\begin{table}[t]
\centering
\caption{Average Jensen–Shannon divergence between the centroid and the draft, and between the centroid and heads, for agree and disagree cases ($c$ denotes the centroid).}
\label{tab:js_agreement}
\small
\begin{tabular}{lcc}
\toprule
 & JS$(c,\text{draft})$ & JS$(c,\text{heads})$ \\
\midrule
Dispersed & 0.0254 & 0.0313 \\
Concentrated & 0.0097 & 0.0014 \\
\bottomrule
\end{tabular}
\end{table}

\begin{algorithm}[h]
  \caption{JS-Divergence-Based Acceptance}
  \label{accept_algo}
  \begin{algorithmic}
    \STATE {\bfseries Input:} draft token $\hat{y}_t$, draft prob $\hat{p}_t$,
    \STATE \hspace{2.8em} LM head probs ${p_t^{(i)}}_{i=1}^K$
    \STATE Compute centroid: $\bar{p}_t = \mathrm{Softmax}\!\left(\frac{1}{K}\sum_{i=1}^{K} l_t^{(i)}\right)$
    \STATE Compute LM head tokens: $y_t^{(i)} = \arg \max p_t^{(i)}$
    \IF{$JS(\hat{p}_t || \bar{p}_t) \leq \max_i JS(p_t^{(i)} || \bar{p}_t )$}
    \STATE Accept
    \ELSIF{$\mathrm{majority}\left(\{ y_t^{(i)} \}_{i=1}^K \right) = \hat{y}_t$}
    \STATE Accept
    \ELSE
    \STATE Reject
    \ENDIF
  \end{algorithmic}
\end{algorithm}

\section{Experiments} \label{Experiments}
In this section, we evaluate the proposed approach, DropMatch, in terms of (i) computational overhead, (ii) acceptance-rate improvements, and (iii) end-to-end decoding speed. We first quantify the cost of MC dropout sampling at the LM head (\cref{sec:4.1}). We then apply the method to standard speculative decoding on Llama-3.1 and Qwen3 model families across GSM8K, MMLU, IFEval, and HumanEval (\cref{sec:4.2}). Finally, we show that the method is complementary to prior lossy and draft model improvements by integrating it with Auto-Judge and EAGLE3 (\cref{sec:4.3}--\cref{sec:4.4}).

\paragraph{Metrics}
We report performance using the following three metrics, which capture key factors affecting speculative decoding.
\begin{itemize}
  \item \textbf{Accuracy}: Unlike lossless speculative decoding, the proposed method does not strictly enforce sampling from the exact target model distribution. To demonstrate that speed improvements are achieved with minimal degradation in task performance, we report accuracy for each benchmark.
  \item \textbf{Mean Acceptance Length ($\tau$)}: This metric measures the average number of draft tokens accepted during verification. Since the proposed method introduces negligible overhead, increases in acceptance length directly translate into inference speedups. Draft length is represented by $L$.
  \item \textbf{Throughput / Speedup}: We measure actual decoding speed in tokens per second (tokens/s) and report relative speedups compared to the baseline model.
\end{itemize}

\subsection{Overhead of MC dropout sampling in the LM head}\label{sec:4.1}
We first demonstrate that applying MC dropout to the LM head incurs only minimal computational overhead. Specifically, we analyze the inference time of the Llama-3.1-70B-Instruct model by decomposing it into transformer blocks and the LM head, and comparing performance with and without MC dropout (\cref{tab:head_overhead}). As observed in recent work~\cite{lu-etal-2025-demystifying}, the computational cost of the LM head is negligible compared to the overall inference cost. Consistent with this observation, our measurements show that the LM head accounts for only 0.05\% of the total forward cost, and that even with five MC dropout paths and JS divergence computation, the total overhead remains at 1.64\%. Based on this observation, we show that selectively using Naive Token-Matching and JS divergence–based criteria allows us to achieve effective speedups while minimizing additional overhead.

\begin{table}[t]
\centering
\caption{Forward latency overhead analysis for the Llama-3.1-70B-Instruct model. Relative computation costs for full forward, LM head forward, MC dropout, and JS divergence, batch size 1 and $n_{input}=5$. Probability for dropout is set to $0.3$}
\label{tab:head_overhead}
\small
\setlength{\tabcolsep}{4pt}
\begin{tabular}{l|ccc}
\toprule
Method & Total (ms) & Head (ms) & Head / Total (\%) \\
\midrule
w/o dropout        & 107 & 0.05 & 0.05 \\
$K=5$ (Naive)        & 104 & 0.17 & 0.17 \\
$K=5$ (JS)   & 106 & 1.74 & 1.64 \\
\bottomrule
\end{tabular}
\end{table}

\subsection{Speculative Decoding with DropMatch} \label{sec:4.2}
For fair comparison, we evaluate speculative decoding with DropMatch using the vLLM~\cite{kwon2023efficient}, lm-evaluation-harness~\cite{eval-harness} and EvalPlus~\cite{evalplus} frameworks. In these experiments, we employ the JS divergence–based acceptance criterion and report results with a batch size of 1 for throughput or speed up. 

\cref{tab:benchmark_comparison} summarizes the experimental results on the Llama-3.1-8B/70B-Instruct and Qwen3 4B/32B models. Both model pairs were evaluated using 4 A100-SXM4-80G GPUs in tensor-parallel mode. Across most tasks, we observe approximately a 10\% increase in acceptance length on draft length $L=5$, which corresponds to a similar improvement in decoding speed over standard speculative decoding. These gains are achieved at almost no additional cost, with little to no degradation in accuracy. Compared to standard speculative decoding, DropMatch achieves a relative speed up to a 1.33× throughput improvement on the Qwen3 4B/32B models at batch size 1 with draft length $L=10$. For the Llama-3.1-8B/70B-Instruct models, DropMatch yields a 1.19× speedup on GSM8K under the same $L=10$ setting. In contrast, for HumanEval, as shown in \cref{tab:agreement_stats}, all heads tend to point to the same token, and the standard model already exhibits a high acceptance rate, making further increases in acceptance length more challenging, particularly due to the strict syntactic requirements of code generation tasks.
\cref{tab:batch_size_speedup} reports relative throughput improvements for the Llama-3.1-70B-Instruct model, as \cref{tab:benchmark_comparison} shown, improvement is consistently maintained at approximately 1.10× even when the batch size is increased to 128.

\begin{table}[h]
\centering
\small
\setlength{\tabcolsep}{5pt}
\caption{Throughput improvement across batch sizes for the GSM8K with draft length $L=10$ on A100-SXM4-80G GPU with Llama3.1-70B-Instruct. Standard speculative decoding(SD) as the baseline (1.0×), both SD and DropMatch(DM) were measured using vLLM with tensor parallelism TP=4.}
\label{tab:batch_size_speedup}
\begin{tabular}{lccccccc}
\toprule
Batch size & 1 & 4 & 16 & 32 & 64 & 128 \\
\midrule
SD + DM & 1.19x & 1.18x & 1.12x & 1.10x & 1.15x & 1.10x \\
\bottomrule
\end{tabular}
\end{table}

\begin{table*}[t]
\centering
\caption{Performance comparison of standard speculative decoding(SD) and its combination with DropMatch(DM) on Llama-3.1 and Qwen3 models. Accuracy, mean acceptance length, and throughput on GSM8K, MMLU, IFEval, and HumanEval benchmarks. Throughput measured with batch size 1. For DropMatch, the dropout probability is set to $p_{drop}=0.3$ with $K=5$ MC dropout paths.}
\label{tab:benchmark_comparison}

\setlength{\tabcolsep}{3.5pt}
\small

\begin{tabular}{ccc|ccc|ccc|ccc|ccc}
\hline
\multirow{2}{*}{Model} & \multirow{2}{*}{Method} & \multirow{2}{*}{$L$}
& \multicolumn{3}{c|}{GSM8K}
& \multicolumn{3}{c|}{MMLU}
& \multicolumn{3}{c|}{IFEval}
& \multicolumn{3}{c}{HumanEval} \\
\cline{4-15}
& &
& Speedup & $\tau$ & Acc.
& Speedup & $\tau$ & Acc.
& Speedup & $\tau$ & Acc.
& Speedup & $\tau$ & Pass@1 \\
\hline
\multirow{4}{*}{\shortstack{Llama-3.1\\[1pt]8B/70B-Instruct}}
% L=5
& SD        & 5  & 1.0x & 4.97 & 94.69 & 1.0x & 4.02 & 85.94 & 1.0x & 4.12 & 85.60 & 1.0x & 5.40 & 81.70 \\
& SD + DM   & 5  & 1.10x & 5.50 & 94.29 & \textbf{1.09x} & 4.43 & 85.50 & \textbf{1.10x} & 4.48 & 85.24 & 1.00x & 5.57 & 79.89 \\
% L=10
& SD        & 10 & 1.02x & 7.62 & 95.40 & 0.86x & 5.25 & 86.06 & 0.87x & 5.47 & 86.86 & \textbf{1.13x} & 8.87 & 81.70 \\
& SD + DM   & 10 & \textbf{1.22x} & 9.20 & 93.90 & 1.02x & 6.06 & 85.34 & 0.99x & 6.34 & 86.17 & 1.12x & 9.38 & 78.70 \\
\hline
\hline
\multirow{4}{*}{Qwen3 4B/32B}
% L=5
& SD        & 5  & 1.0x & 5.06 & 86.20  & 1.0x & 4.62 & 83.21  & 1.0x & 3.40 & 80.72  & 1.0x & 5.09 & 90.4 \\
& SD + DM   & 5  & 1.07x & 5.66 & 87.49  & 1.05x & 5.15 & 82.15  & \textbf{1.19x} & 4.22 & 78.24 & 1.00x & 5.27 & 90.9 \\
% L=10
& SD        & 10 & 0.96x & 7.87 & 86.20 & 1.12x & 6.48 & 83.26  & 0.72x & 4.04 & 82.31  & 1.31x & 8.11 & 92.1 \\
& SD + DM   & 10 & \textbf{1.13x} & 9.78 & 86.58  & \textbf{1.49x} & 8.60 & 82.06  & 0.98x & 5.67 & 81.60  & \textbf{1.39x} & 8.72 & 90.2 \\
\hline
\end{tabular}
\end{table*}

\begin{table*}[ht]
\centering
\caption{Performance comparison of EAGLE3 and EAGLE3 + DropMatch(DM) on GSM8K, MT-bench, Alpaca with the Llama-3.3-70B-Instruct EAGLE3 model across different draft lengths. For our method, the dropout probability is set to $p_{drop}=0.3$ with $K=5$ MC dropout paths. Speedups are reported relative to the standalone baseline (1.0×).}
\label{tab:eagle3_comparison}

\setlength{\tabcolsep}{6pt}
\small

\begin{tabular}{
>{\centering\arraybackslash}p{2.0cm}
>{\centering\arraybackslash}p{1.3cm}
>{\centering\arraybackslash}p{0.3cm}
| c c c | c c c | c c c
}
\hline
\multirow{2}{*}{Model} & \multirow{2}{*}{Method} & \multirow{2}{*}{$L$}
& \multicolumn{3}{c|}{GSM8K}
& \multicolumn{3}{c|}{MT-bench}
& \multicolumn{3}{c}{Alpaca} \\
\cline{4-12}
& & 
& Speedup & $\tau$ & Acc.
& Speedup & $\tau$ & Score
& Speedup & $\tau$ & Win rate \\
\hline

% ---------- L33 70B ----------
\multirow{3}{*}{\shortstack{Llama-3.3\\[1pt]70B-Instruct}}
& \multirow{3}{*}{EAGLE3}
& 5 & 4.19x & 5.82 & 78.75 & 3.90x & 5.37 & 8.47 & 4.31x & 5.83 & 50.0 \\
& & 7 & 4.65x & 6.71 & 78.75 & 4.18x & 5.81 & 8.43 & 4.75x & 6.59 & 51.3 \\
& & 9 & 4.68x & 6.71 & 78.75 & 4.24x & 6.06 & 8.48 & 4.92x & 6.89 & 50.6 \\
\hline

% ---------- L33 8B/70B ----------
\multirow{3}{*}{\shortstack{Llama-3.3\\[1pt]70B-Instruct}}
& \multirow{3}{*}{\shortstack[r]{EAGLE3\\[1pt]+DM}}
& 5 & 4.32x & 6.09 & 78.75 & 4.13x & 5.86 & 8.45 & 4.47x & 6.30 & 55.0 \\
& & 7 & 4.86x & 7.07 & 81.25 & 4.52x & 6.58 & 8.46 & 5.02x & 7.25 & 56.0 \\
& & 9 & \textbf{5.03x} & \textbf{7.48} & 80.00 & \textbf{4.62x} & \textbf{6.89} & 8.59 & \textbf{5.27x} & \textbf{7.74} & 52.0 \\
\hline

\end{tabular}
\end{table*}

\subsection{Auto-Judge with DropMatch} \label{sec:4.3}
\begin{figure}[t]
  \vskip 0.2in
  \begin{center}
    \centerline{\includegraphics[width=\columnwidth]{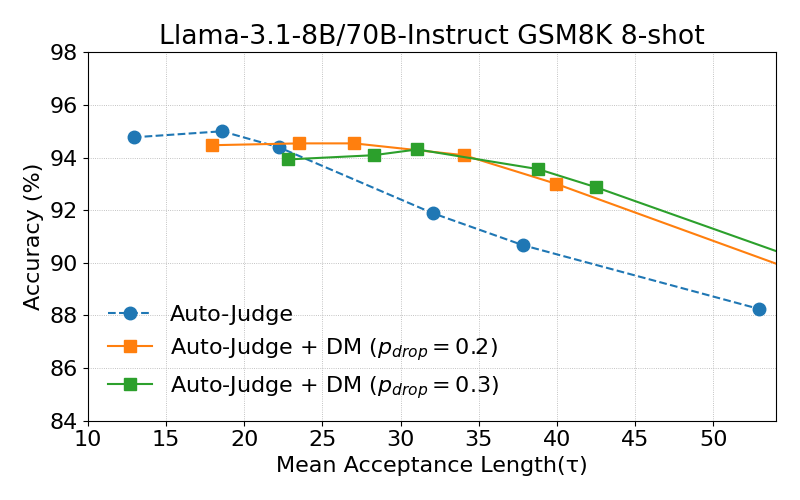}}
    \caption{Comparison of Auto-Judge and Auto-Judge combined with DropMatch(DM) on GSM8K 8shot with Llama-3.1-8B/70B-Instruct models. Accuracy and mean acceptance length graphs at dropout probabilities $p_{drop}=0.2$ and $0.3$ with $K=5$ MC dropout paths. A rightward shift of Auto-Judge + DM relative to Auto-Judge indicates increased acceptance length at comparable accuracy levels.}
    \label{fig:auto_judge_gsm8k_8shots}
  \end{center}
\end{figure}

Following \cref{sec:4.1}, we apply DropMatch to the Auto-Judge framework. In this experiment, we use the Llama-3.1-70B-Instruct model and evaluate performance on the GSM8K and LiveCodeBench~\cite{jain2024livecodebench} datasets, following the experimental setup of Auto-Judge. We reproduce the Auto-Judge experiments by training judge heads using the provided training datasets and the official repository\footnote{https://github.com/garipovroma/autojudge} code for both GSM8K and LiveCodeBench.
\cref{fig:auto_judge_gsm8k_8shots} shows that on GSM8K, under identical Auto-Judge parameter settings, the proposed method maintains comparable accuracy while achieving longer mean acceptance length than the baseline Auto-Judge approach. \cref{tab:auto_judge_gsm8k_8shots} reports throughput comparisons with batch size set to 1, demonstrating that our method provides effective speedups under Auto-Judge inference setup. Across all threshold configurations, the proposed approach consistently achieves longer acceptance length than the baseline while minimizing performance degradation and improving decoding speed from 1.06x to 1.29x compared to Auto-Judge and 1.44× to 2.11× compared to the standard model.
\cref{tab:auto_judge_threshold} summarizes results on the LiveCodeBench dataset, reporting mean acceptance length and accuracy across different judge-head threshold settings. On this dataset as well, DropMatch consistently increases acceptance length across all parameter configurations, highlighting its robustness when combined with Auto-Judge.

\begin{table}[t]
\centering
\caption{Performance comparison of Auto-Judge and Auto-Judge combined with DropMatch (DM) on Llama-3.1-8B/70B-Instruct with GSM8K 8-shot. Accuracy and throughput (batch size 1) are reported across thresholds. Standard speculative decoding (SD) achieves 66.6 and 45.5 tokens/s with $L=8$ and $L=32$, respectively. Auto-Judge uses $L=32$, and DropMatch uses $p_{drop}=0.3$ with $K=5$.}
\label{tab:auto_judge_gsm8k_8shots}

\small
\setlength{\tabcolsep}{2pt}

\begin{tabular}{
>{\centering\arraybackslash}m{2.4cm}
| >{\centering\arraybackslash}m{2.0cm}
| c c c c
}
\hline
\multirow{2}{*}{\textbf{Method}} &
\multirow{2}{*}{\textbf{Metric}} &
\multicolumn{4}{c}{\textbf{Threshold}} \\
\cline{3-6}
 & & 0.014 & 0.022 & 0.112 & 0.281 \\
\hline

\multirow{2}{*}{Auto-Judge}
& Accuracy, \% & 95.1 & 94.3 & 91.4 & 88.2 \\
& tokens / s   & 57.4 & 62.2 & 84.8 & 90.1 \\
\hline

\multirow{2}{*}{Auto-Judge + DM}
& Accuracy, \% & 94.1 & 93.9 & 93.0 & \textbf{90.7} \\
& tokens / s   & 73.9 & 77.6 & 90.4 & \textbf{95.9} \\
\hline

\multirow{3}{*}{Speedup(DM)}
& Auto-Judge=1x & 1.29x & 1.21x & 1.07x & \textbf{1.06x} \\
& SD($L{=}8$)=1x  & 1.11x & 1.16x & 1.36x & \textbf{1.44x} \\
& SD($L{=}32$)=1x & 1.62x & 1.71x & 1.99x & \textbf{2.11x} \\
\hline
\end{tabular}
\end{table}

\begin{table}[h]
\centering
\caption{Pass@1 and mean acceptance length comparison of Auto-Judge and Auto-Judge + DropMatch(DM) on LiveCodeBench with the Llama-3.1-70B-Instruct model across judge-head thresholds. Dropout probability $p_{drop}=0.3$ and number of MC dropout paths $K=5$ for DropMatch.}
\label{tab:auto_judge_threshold}

\small
\setlength{\tabcolsep}{6pt}

\begin{tabular}{c|cc|cc}
\hline
\multirow{2}{*}{Threshold}
& \multicolumn{2}{c|}{Auto-Judge}
& \multicolumn{2}{c}{Auto-Judge + DM} \\
\cline{2-5}
& $\tau$ & Pass@1 & $\tau$ & Pass@1 \\
\hline
0.0004 & 12.037 & 31.14 & 16.229 & 29.32 \\
0.0123 & 14.141 & 29.09 & 18.587 & 29.43 \\
0.0255 & 18.985 & 27.95 & 24.025 & 29.09 \\
0.0508 & 27.549 & 28.07 & 32.674 & 26.82 \\
0.1214 & 46.974 & 23.98 & 49.481 & 22.39 \\
\hline
\end{tabular}
\end{table}

\subsection{EAGLE3 with DropMatch} \label{sec:4.4}
For experiments with EAGLE3, we adopt the Naive Token-Matching criterion instead of the JS divergence–based acceptance rule. EAGLE3 models employ tree decoding, where candidate tokens are sampled from the same probability distribution, making JS divergence–based evaluation redundant and computationally inefficient due to repeated calculations. In contrast, the Naive Token-Matching criterion introduces no additional divergence computation overhead and can be applied even when the vocabularies of the draft and target models differ, enabling broader applicability. In addition, to ensure faithful reproduction of EAGLE3, we conduct all experiments using the official EAGLE3 repository\footnote{https://github.com/SafeAILab/EAGLE} code.

\cref{tab:eagle3_comparison} presents results for Llama-3.3-70B-Instruct EAGLE3 model combined with the proposed method, showing that acceptance length and decoding speed improvements are preserved. While increasing draft length in EAGLE3 alone eventually leads to saturation in acceptance length and speed gains, combining it with our method enables additional acceleration without significantly degrading performance. The experiments in \cref{tab:eagle3_comparison} follow the EAGLE3 setting, evaluating 80 samples with various draft lengths. GSM8K provides ground-truth answers, and performance is evaluated based on exact answer correctness. In contrast, MT-Bench~\cite{zheng2023judgingllmasajudgemtbenchchatbot} and Alpaca~\cite{alpaca} follow an LLM-as-a-Judge~\cite{li2025whosjudgedetectabilityllmgenerated} evaluation paradigm to assess response quality. For MT-Bench, scores are reported on a 10 point scale, while for Alpaca, performance is measured using win rates by comparing the responses of EAGLE3 and EAGLE3+DropMatch against those produced by standard speculative decoding.
When evaluating accuracy on the full GSM8K dataset, the standard model and EAGLE3 achieves 81.7\%, while the model combining EAGLE3 with DropMatch attains 81.3\% with $L=10$, indicating that performance is largely preserved. For MT-Bench and Alpaca datasets, evaluation is conducted using GPT-4~\cite{openai2024gpt4technicalreport} as the judge model; under this setting, EAGLE3 model scores 8.64 and win rate 50\% against standard on the full datasets, respectively.
Applying DropMatch to EAGLE3 yields scores of 8.52 and win rate 48\% against standard with $L=10$, demonstrating that additional acceleration can be obtained with minimal degradation in quality assessment. Across all experimental settings, combining EAGLE3 with DropMatch yields consistent speed improvements and effectively extends acceptance length. Notably, on GSM8K, where EAGLE3 alone saturates beyond a certain draft length, our approach allows acceptance length to be further increased while largely preserving task performance.

\begin{figure}[H]
  \vskip 0.2in
  \begin{center}
    \centerline{\includegraphics[width=\columnwidth]{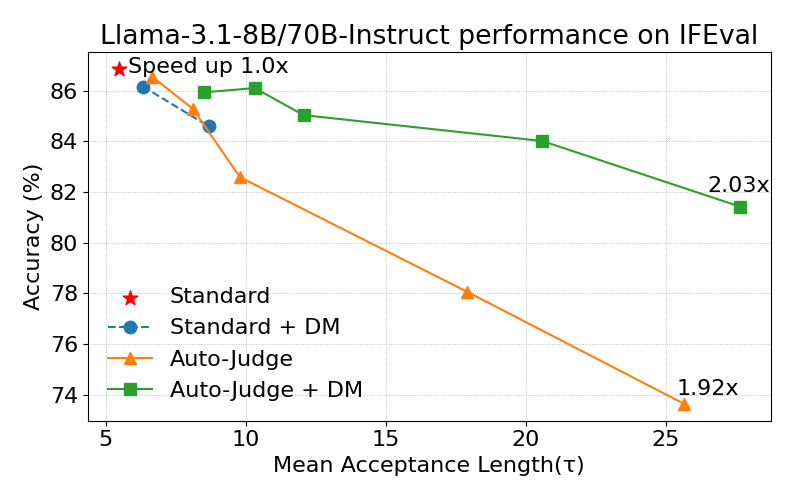}}
    \caption{
      Performance of Auto-Judge and Auto-Judge combined with DropMatch(DM) on IFEval with Llama-3.1-8B/70B-Instruct models. Accuracy and mean acceptance length graphs at dropout probability $p_{drop}=0.3$ with $K=5$ multiple paths. Auto-Judge exhibiting increased decoding speed with longer acceptance length, similar to \cref{tab:auto_judge_gsm8k_8shots}, but showing rapid performance degradation under out-of-distribution conditions.
    }
    \label{fig:auto_judge_ifeval}
  \end{center}
\end{figure}

\subsection{Out of Distribution Performance}
Finally, we evaluate learning-based speculative decoding methods and our proposed approach on out-of-distribution (OOD) data, i.e., data that the methods have not been trained on. Since our approach does not modify the parameters of either the draft or the target model, it avoids catastrophic forgetting and mitigates degradation under distribution shifts. \cref{fig:auto_judge_ifeval} reports results on the IFEval benchmark using Auto-Judge, whose judge head is trained on mathematical data as described in \cref{sec:4.3}. Although Auto-Judge continues to achieve a substantially longer mean acceptance length than standard speculative decoding, its task performance degrades rapidly as the distribution shifts. In contrast, our method maintains stable performance even when the draft length is increased, while avoiding excessively high acceptance rates. Moreover, when combined with Auto-Judge, DropMatch achieves a longer mean acceptance length under the same experimental settings, while exhibiting a more gradual degradation in performance. This trend is also observed in Fig.~\ref{fig:auto_judge_gsm8k_8shots}, suggesting that Auto-Judge frequently rejects tokens that should be accepted to improve performance, or conversely accepts tokens that should be rejected in out-of-distribution cases. By complementing these failure modes, DropMatch effectively extends acceptance length while mitigating performance degradation.

\begin{table}[h]
\centering
\caption{Performance of EAGLE3 and standard speculative decoding + DropMatch(DM) on KoMT-bench with draft length set to $L=7$, using Llama-3.3-70B-Instruct as the target model and Llama-3.1-8B-Instruct as the draft model.}
\label{tab:eagle3_komt}

\small
\setlength{\tabcolsep}{6pt}

\begin{tabular}{lcc}
\hline
Model & Score & $\tau$ \\
\hline
Standard        & 8.08 & 4.85 \\
EAGLE3    & 7.96 & 1.46 \\
Standard + DM   & 8.12 & 5.24 \\
\hline
\end{tabular}
\end{table}

\cref{tab:eagle3_komt} presents results on the KoMT-bench benchmark, which consists of Korean translation data, using EAGLE3—a draft model trained on English data. These results indicate that, in the case of EAGLE3, the target model has difficulty accepting tokens proposed by the draft model when a distribution shift occurs. This tendency persists even when the draft length is increased from $L=5$(Fig.~\ref{fig:komt_ood:b}) to $L=7$. In contrast, our method remains effective, demonstrating strong adaptability to data distributions encountered during pretraining. Overall, these results suggest that our approach effectively avoids out-of-distribution scenarios and can be readily applied in an off-the-shelf manner, without additional training or parameter updates.

\section{Conclusion}
In this work, we propose DropMatch, a novel approach to accelerate speculative decoding by applying Monte Carlo (MC) dropout exclusively to the LM head of the target model for sampling-based acceptance decisions. The proposed method is training-free, making it less susceptible to out-of-distribution issues during acceptance judgment, and operates in a data-free and calibration-free manner. Since it requires no modification to the architecture of pretrained models and can be easily applied by introducing MC dropout only at the LM head, the computational overhead is negligible. Experimental results demonstrate that our method achieves inference speedups ranging from 1.09× to 1.33× compared to standard speculative decoding. Furthermore, the proposed approach is not limited to standard speculative decoding and can be seamlessly combined with acceleration methods that require additional architectures or training, consistently yielding further speed improvements across diverse settings.

% In the unusual situation where you want a paper to appear in the
% references without citing it in the main text, use \nocite
% \nocite{langley00}

\bibliography{references}
\bibliographystyle{icml2026}

\end{document}